%% file: main_arxiv.tex
\documentclass[letterpaper]{article} 
\usepackage{aaai25}  
\usepackage{times}  
\usepackage{helvet}  
\usepackage{courier}  
\usepackage[hyphens]{url}  
\usepackage{graphicx} 
\usepackage{natbib}  
\usepackage{caption} 
\usepackage{algorithm}
\usepackage{arydshln}
\usepackage{tabularx}
\usepackage{ragged2e}
\usepackage{algorithmic}

\usepackage{comment}
\usepackage{multirow}
\usepackage{amsmath}
\usepackage[inline]{enumitem}
\usepackage{newfloat}
\usepackage{listings}
\DeclareCaptionStyle{ruled}{labelfont=normalfont,labelsep=colon,strut=off} 
\floatstyle{ruled}
\newfloat{listing}{tb}{lst}{}
\floatname{listing}{Listing}
\usepackage{xcolor}
\definecolor{ForestGreen}{rgb}{0.0, 0.8, 0.0}
\usepackage{booktabs,colortbl}

\title{uMedSum: A Unified Framework for Advancing Medical Abstractive Summarization}
\author{
    Aishik Nagar\textsuperscript{\rm 1}, 
    Yutong Liu\textsuperscript{\rm 1},
    Andy T. Liu\textsuperscript{\rm 1},
    Viktor Schlegel\textsuperscript{\rm 2}\thanks{Work done as part of ASUS Intelligent Cloud Services (AICS)},
    Vijay Prakash Dwivedi\textsuperscript{\rm 1},
    Arun-Kumar Kaliya-Perumal\textsuperscript{\rm 3}, 
    Guna Pratheep Kalanchiam\textsuperscript{\rm 4},
    Yili Tang\textsuperscript{\rm 1},
    Robby T. Tan\textsuperscript{\rm 1}
}
\affiliations{
    \textsuperscript{\rm 1}ASUS Intelligent Cloud Services (AICS), Singapore, 
    \textsuperscript{\rm 2}Imperial College London, Imperial Global Singapore, 
    \textsuperscript{\rm 3}Rehabilitation Research Institute of Singapore, Lee Kong Chian School of Medicine, Nanyang Technological University, Singapore, 
    \textsuperscript{\rm 4}Division of Spine, Department of Orthopaedic Surgery, Tan Tock Seng Hospital, Singapore\\   
    
    \texttt{\{aishik\_nagar; yutong\_liu\}@asus.com}
}

\usepackage{bibentry}

\newcommand{\frameworkname}{\texttt{uMedSum}}

\begin{document}

\maketitle

\begin{abstract}
Medical abstractive summarization faces the challenge of balancing faithfulness and informativeness. Current methods often sacrifice key information for faithfulness or introduce confabulations when prioritizing informativeness. While recent advancements in techniques like in-context learning (ICL) and fine-tuning have improved medical summarization, they often overlook crucial aspects such as faithfulness and informativeness without considering advanced methods like model reasoning and self-improvement. Moreover, the field lacks a unified benchmark, hindering systematic evaluation due to varied metrics and datasets. This paper addresses these gaps by presenting a comprehensive benchmark of six advanced abstractive summarization methods across three diverse datasets using five standardized metrics. Building on these findings, we propose \frameworkname{}, a modular hybrid summarization framework that introduces novel approaches for sequential confabulation removal followed by key missing information addition, ensuring both faithfulness and informativeness. Our work improves upon previous GPT-4-based state-of-the-art (SOTA) medical summarization methods, significantly outperforming them in both quantitative metrics and qualitative domain expert evaluations. Notably, we achieve an average relative performance improvement of 11.8\% in reference-free metrics over the previous SOTA. Doctors prefer \frameworkname{}'s summaries 6 times more than previous SOTA in difficult cases where there are chances of confabulations or missing information. These results highlight \frameworkname{}'s effectiveness and generalizability across various datasets and metrics, marking a significant advancement in medical summarization. 
\end{abstract}

\section{Introduction}
Large Language Models (LLMs) have shown exceptional performance in generative tasks, including zero-shot and out-of-the-box applications in specialized areas like summarization~\cite{lei2023chain, van2024adapted}. In the medical field, document summarization holds promise for greatly improving the efficiency of medical staff in reviewing lengthy documents, such as medical exam reports or patient histories.
However, the stochastic nature of LLMs and their lack of formal guarantees~\cite{Li2023Team:PULSARModels,Schlegel2023PULSARRecords} often lead to summaries that deviate from input documents, limiting their practical usability.

This is particularly problematic in the medical domain, where accurate and complete information is crucial for effective decision-making. Doctors rely on summaries that capture all relevant details without introducing erroneous information, emphasizing two critical aspects of medical summarization: \emph{faithfulness} and \emph{informativeness}.
Lack of faithfulness results in confabulations, where parts of the summary contain information not present in the input document~\cite{maynez-etal-2020-faithfulness}.
Insufficient informativeness leads to omitting relevant details from the input document~\cite{mao2020constrained}.
Such summaries can provide doctors with incomplete evidence or inaccurate information, potentially leading to misdiagnoses or inappropriate treatment decisions, ultimately impacting patient outcomes.

\begin{figure*}[ht]
    \centering
    \includegraphics[width=1\linewidth]{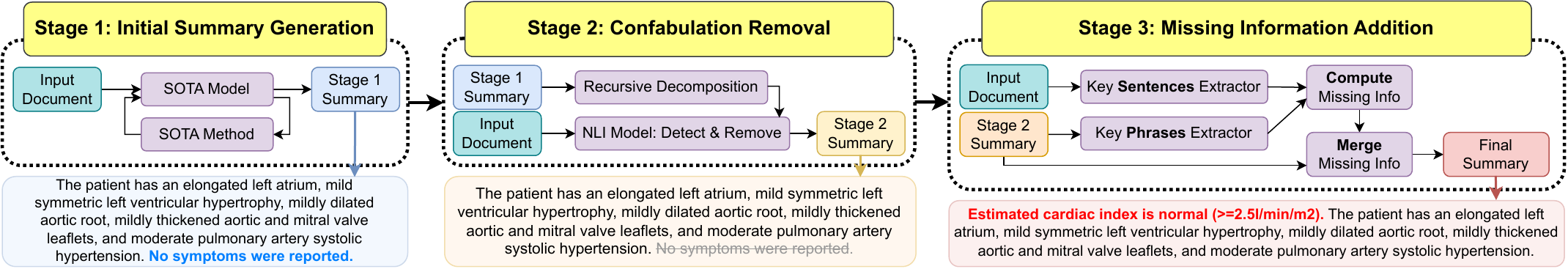}
    \caption{Overview of the proposed three-stage framework.
    The process is illustrated with example outputs at each stage when using \frameworkname{} with Element Aware Summarization and GPT-4. 
    {\color{blue}Blue text} indicates confabulated information (information not grounded in the input document), while {\color{red}red text} highlights added key information previously missing from the summary.}
    \label{fig:main-framework}
\end{figure*}

\begin{figure}[!t]
    \centering
    \includegraphics[width=1\linewidth]{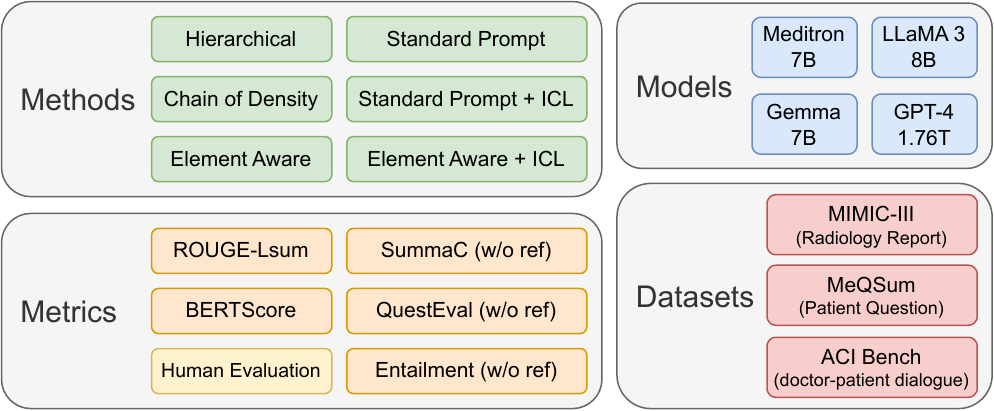}
    \caption{Overview of the proposed medical summarization benchmark for fair comparison.
    }
    \label{fig:benchmark}
\end{figure}

Current efforts in enhancing faithfulness and informativeness for summarization face several limitations:
\begin{enumerate*}[label=\emph{(\roman*)}]
\item  Many techniques address only specific sub-problems (e.g., faithfulness or informativeness) in isolation;
\item  Most approaches focus on reference-based confabulation detection, while reference-free confabulation removal in summarization remains an open problem. Moreover, overzealous removal of confabulated content may lead to summaries with missing information;
\item Existing methods often rely on either purely abstractive or extractive techniques, without leveraging both; and
\item Hybrid exceptions to the previous point, such as Constrained Abstractive Summarization (CAS)~\cite{mao2020constrained}, which aim to add missing information into abstractive summaries, inherit limitations from their constituent parts. Specifically,  confabulations in the initial abstractive summary will persist in the final summary.
\end{enumerate*}

Advancements in medical summarization are further impeded by the lack of standardized benchmarks, driven by inconsistent metric and dataset choices in prior studies, and insufficient evaluations of faithfulness and informativeness. Recent approaches rely on task adaptation such as in-context learning (ICL)~\cite{van2024adapted} and parameter-efficient fine-tuning like QLoRA~\cite{dettmers2023qlora}, but neglect recent model reasoning advancements \cite{chang2023booookscore}. Secondly, by mainly showing improvement on reference-based metrics, they neglect important summary aspects not captured by those, such as faithfulness and informativeness~\cite{maynez-etal-2020-faithfulness}.

In light of these considerations, we present a comprehensive medical summarization benchmark and a large-scale study of previous SOTA summarization methods. 
We introduce a unified summarization framework, \frameworkname{}, which tackles these issues through a hybrid abstractive and extractive approach, including novel techniques for removing confabulated information as well as adding missing information to balance faithfulness and informativeness. 
By sequentially removing confabulated information followed by adding missing information, \frameworkname{} avoids the issue of overzealous removal of information during faithfulness checks, while ensuring missing information is not added to unfaithful summaries. 
We then combine the benchmark findings with \frameworkname{} to outperform the existing SOTA. 
Our contributions include:
\begin{enumerate}
\item A comprehensive benchmark comparing six recent summarization methods across three diverse datasets using five standardized metrics, including both reference-based and reference-free metrics, and human evaluation by clinical doctors (Figure~\ref{fig:benchmark}).
\item A novel three-stage framework, \frameworkname{} (Figure~\ref{fig:main-framework}), that enhances current summarization method and model:
\begin{enumerate}
\item Initial summary generation using the best-performing method-LLM combination from our benchmark.
\item NLI-based confabulation detection and removal.
\item Hybrid abstractive-extractive approach for incorporating missing key information.
\end{enumerate}
\item We achieve significant improvements in faithfulness and informativeness compared to previous GPT-4-based SOTA methods~\cite{van2024adapted} while maintaining competitive performance on reference-based metrics.
\item Human evaluation shows that domain experts prefer our framework's summary 6 times more than previous SOTA in difficult cases, with equal preference in straightforward cases.
\item An open-source benchmarking toolkit and code to facilitate further research in medical summarization.
\end{enumerate}

Our work takes a significant step toward more reliable and clinically applicable text summarization systems by addressing the crucial aspects of faithfulness and informativeness in medical summarization.

\section{Related Work}

Summarization is usually approached by extractive and abstractive approaches~\cite{nenkova2011automatic,luo2024comprehensive}. 
\textbf{Extractive summarization} selects key sentences or phrases directly from the input document. For example, more recent extractive summarization approaches explore semantic matching \citep{zhong2020extractive}, domain-specific term extraction using BERT embeddings \citep{sammet2023domain}, and advancements in keyphrase extraction with pre-trained language models \citep{song2023survey}.
\textbf{Abstractive summarization}, conversely, aims to rephrase content for more concise and readable summaries. More recent advancements over traditional sequence-to-sequence approaches~\cite{lewis-etal-2020-bart,zhang2020pegasus} include element-aware steering of summary content~\citep{wang2023element}, coherence assessment for long documents~\citep{chang2023booookscore}, and reinforcement learning for factual consistency~\citep{roit2023factually}. 
Notably, most existing work applies either extractive or abstractive techniques in isolation, potentially limiting their effectiveness. Furthermore, challenges remain with confabulations~\citep{maynez-etal-2020-faithfulness} and incompleteness~\cite{mao2020constrained}.

\paragraph{Confabulated Information Detection.} Confabulation detection, or identifying information not grounded in the input document, is a major challenge in medical summarization. While existing methods typically provide a general factuality score for generated summaries \cite{maynez-etal-2020-faithfulness, liu2024hit, ji2023towards}, they often fail to systematically remove confabulated information. Some techniques attempt to address epistemic uncertainty by leveraging internal logit-level data \cite{yadkori2024believe, manakul2023selfcheckgpt, chen2024inside, farquhar2024detecting}, but these are generally limited to question-answering scenarios with clear ground truths and require access to internal model states, making them less scalable and compatible with proprietary models. Inspired by \citet{maynez-etal-2020-faithfulness} and \citet{lei2023chain}, our framework uses entailment-based metrics for removing confabulated spans in summaries, employing sentences and atomic facts for more accurate removal \cite{thirukovalluru2024atomic}. We also address the common issue of overzealous removal by detecting and reintegrating key missing information afterward.

\paragraph{Missing Information Addition.}
Missing information in summaries is addressed by extractive or hybrid techniques which rely on extractions to identify key missing details. Despite improving key phrase identification, extractive summaries often struggle with verbosity and low fluency \cite{nathan2023investigating}. Hybrid approaches like Constrained Abstractive Summarization (CAS)~\cite{mao2020constrained} address missing information but do not consider confabulations in the generated summary. 
In contrast, our work proposes a novel approach that effectively combines extractive and abstractive methods to address both confabulations and key missing information. 
We show both quantitatively and qualitatively, that this integration achieves a delicate balance, leveraging both approaches' strengths while mitigating their weaknesses.

\section{\frameworkname{}: Faithfulness and Informativeness in Medical Summarization}

\paragraph{Summarization Benchmark.}To address the lack of a systematic benchmark for summarization methods in medical summarization, we first evaluate four recent methods: Standard Prompting (baseline), Element-Aware Summarization with Large Language Models \cite{wang2023element}, Chain of Density \cite{addams2023sparse}, and Hierarchical Summarization \cite{chang2023booookscore}. 
Each technique offers distinct benefits and drawbacks. For instance, Element-Aware Summarization enhances content relevance by targeting domain-specific elements, while Chain of Density produces information-dense but less readable summaries. Hierarchical Summarization effectively addresses the “lost-in-the-middle” issue in long contexts \cite{ravaut2023context}. We then combine the top-performing methods with task adaptation strategies, particularly In-Context Learning, which outperforms QLoRA for similar tasks \cite{van2024adapted}. This step ensures the highest possible quality for the initial summary, laying a strong foundation for \frameworkname{}. 

\paragraph{\frameworkname{}.}The \frameworkname{} pipeline is designed to produce high-quality, faithful, and comprehensive medical summaries through a three-stage process, visualized in Figure \ref{fig:main-framework}: Initial Summary Generation, Confabulation Removal, and Missing Information Addition. 
Each stage of \frameworkname{} is a modular, self-contained component that can be independently updated and tuned, providing flexibility and efficiency with minimal computational overhead. This modularity allows for seamless integration with both open and closed-source models during inference.

\subsection{Stage 1: Initial Summary Generation}

The combination of best-performing methods and models from the benchmark is selected for further evaluation and enhancement using the stages of \frameworkname{}. In the first stage, we generate an initial abstractive summary given the input document.

\subsection{Stage 2: Confabulation Removal (Faithfulness)}

The \frameworkname{} pipeline takes a novel approach by repurposing Natural Language Inference (NLI) models to not just evaluate the factuality of generated summaries but to directly detect and remove discrete confabulated information from generated summaries. This approach differs from existing methods, which typically focus on sentence-level entailment or entity-based splitting \cite{lei2023chain}, by introducing a more granular decomposition based on atomic facts~\cite{thirukovalluru2024atomic, nawrath2024role, stacey2023logical}.
Specifically, we propose a two-step process: \emph{(1)} summary decomposition into smaller, manageable units, and \emph{(2)} pairwise NLI-based confabulation detection and removal. 

\paragraph{Summary Decomposition.} We begin by decomposing the summary generated in Stage~1 into smaller units called Summary Content Units (SCUs)~\citep{nawrath2024role} or Decomposed Summary Units~(DSUs). We propose \textit{Recursive Threshold-based Text Segmentation} to further split sentences into clause-level atomic facts. Unlike previous works that stop at sentence-level decomposition or rely on entity-based splitting for further decomposition \cite{lei2023chain}, our approach aims to create self-contained units that encapsulate atomic facts. Atomic facts encapsulate the smallest meaningful statements that can stand alone as true or false propositions. This aligns well with the NLI task, where the goal is to determine the logical relationship (entailment, contradiction, or neutrality) between two statements. This atomic view of facts allows us to detect confabulations precisely.

Formally, let \( D_k \) represent a decomposed summary unit (DSU) from summary \( S_i \), where \( k \) indexes the specific unit. The NLI model computes the entailment score \( E(D_k) \) for each DSU, where the score \( E(D_k) \) represents the probability distribution over entailment labels (entailment, neutral, contradiction).

\paragraph{Recursive Threshold-Based Text Segmentation (RTB-TS).} 

The algorithmic details are described as follows:
\begin{itemize}
\item Initial Segmentation:
We begin by decomposing the summary \( S_i \) into DSUs \( D_k \) using a sentence boundary disambiguation technique:
$S_i \rightarrow \{D_1, D_2, \dots, D_k\}$.
This initial step provides a coarse segmentation based on sentence boundaries.

\item Pairwise NLI Scoring:
For each DSU \( D_k \), we compute the entailment score \( E(D_k) \) using a fine-tuned NLI model:
\begin{eqnarray}
E(D_k) &=& P(\text{entailment} \mid I, D_k),\\
N(D_k) &=& P(\text{neutral} \mid I, D_k),\\
C(D_k) &=& P(\text{contradiction} \mid I,  D_k).
\end{eqnarray}

\item Thresholding and Segmentation:
We apply a threshold to the entailment score to find the initial classification:
\begin{eqnarray}
\text{Class}_\text{Entailed}(D_k) &:& E(D_k) > T_e,\\
\text{Class}_\text{Confab}(D_k) &:& N(D_k) + C(D_k) > T_c,\\
\text{Class}_\text{Uncertain}(D_k) &:& \text{otherwise},
\end{eqnarray}
where \( T_e \) is the entailment threshold and \( T_c \) is the confabulation threshold.
If \( D_k \) is classified as "uncertain," further segmentation is necessary.

\noindent

\item Recursive Decomposition:
For DSUs in the "uncertain" category, we recursively apply segmentation based on the identification of atomic facts within the DSU. This involves breaking down $D_k$ into finer sub-units $D_{k,a}$ where $a$ indexes each atomic fact:
$
D_k \rightarrow \{D_{k,1}, D_{k,2}, \dots, D_{k,a}\}
$.
We recompute the entailment score for each atomic fact sub-unit $D_{k,a}$:
\begin{eqnarray}
E(D_{k,a}) = P(\text{entailment} \mid I, D_{k,a}),
\end{eqnarray}
and retain only those atomic facts where the value of $E(D_{k,a})$ is greater than a chosen threshold.

\item Aggregation of Faithful DSUs:
After recursive segmentation and filtering, we concatenate ($\oplus$) the remaining faithful DSUs to form the refined summary:
\begin{eqnarray}
S_i^{\text{refined}} = \bigoplus_k D_k^{\text{faithful}}.
\end{eqnarray}
The final refined summary \( S_i^{\text{refined}} \) has suppressed the confabulated atomic facts according to the NLI model.
\end{itemize}

\subsection{Stage 3: Missing Information Addition (Informativeness)}
Hybrid methods like \cite{mao2020constrained} risk confabulated initial summaries. Our approach separates confabulation removal (Stage 2) before adding missing key information (Stage 3), reducing the chance of new confabulations in the final summary.
To capture key information from the input document, we identify key sentences in the document and key phrases in the Stage 2 summary. We introduce a novel approach to measure coverage of key information in the summary and integrate missing information into the appropriate sections of the summary to maintain consistency and readability.

\paragraph{Key Information Extraction.}
Our extracted key information from either the input document or the summary is described as follows. Let $K_{\rm doc} = \{ k_{\rm doc}^i \mid i \leq \rm top_M \}$ for the source document, and $K_{\rm summ} = \{ k_{\rm summ}^i \mid i \leq \rm top_N \}$ for the generated summary. Where $K_{\rm doc}$ are key sentences from the input document; $K_{\rm summ}$ are key phrases from our summary; $\rm top_M$ and $\rm top_N$ are the thresholds for the number of key sentences and key phrases, respectively.

For the input document, we use sentences as the minimum unit of granularity for extraction. For the summary generated in Stage 2, we apply a key phrase extraction method, such as the one used by~\citet{grootendorst2020keybert}, which extracts $n$-grams as key phrases. We then iteratively rank the sentences or phrases using $\rm MMR$~\cite{bennani2018simple} and select the top-$K$ as key sentences or key phrases. The complete algorithm for this process is described in the Appendix.

\paragraph{Missing Key Information Detection.}

Given $K_{\rm doc}$ and $K_{\rm summ}$ extracted from the input document and generated summary respectively, we calculate coverage scores ${\rm cov}_{\rm score}^{i}$ for each $k_{\rm doc}^i$ based on $K_{\rm summ}$. 
Specifically, we compute the embedding matrices for key sentences and key phrases \cite{reimers-2019-sentence-bert}. Let $\text{Embed}_{\rm doc}$ represent the matrix formed by stacking the embeddings of the key sentences from the input document, and $\text{Embed}_{\rm summ}$ represent the matrix formed by stacking the embeddings of the key phrases from the Stage 2 summary. Here, $\text{Embed}_{\rm doc}$ is of size $m \times d$, where $m$ is the number of key sentences in the input document, and $d$ is the embedding dimension. Similarly, $\text{Embed}_{\rm summ}$ is of size $n \times d$, where $n$ is the number of key phrases in the Stage 2 summary.

The similarities between \( K_{\rm doc} \) and \( K_{\rm summ} \) are computed as the dot product of the document and summary embedding matrices, yielding a similarity matrix \( [{\rm sim}^{i, j}]_{m \times n} \). Coverage scores for key sentences in the document are then determined by taking the maximum similarity for each sentence across the key phrases from summary, resulting in a vector \( \text{Cov}_{\rm score} \) of size \( m \times 1 \).

We define the coverage score of the $i$-th sentence as \( {\rm cov}^i_{\rm score} = \max_{ j \leq n} \{{\rm sim}^{i, j}\} \) and introduce a threshold parameter \( {\rm cov}_{\rm min} \). Any \( k_{\rm doc}^{i} \) with a coverage score below \( {\rm cov}_{\rm min} \) is considered missing information. The set of potential missing information is represented as:
\[
K_{\rm missing} = \{k_{\rm doc}^{i} \mid i \leq m, {\rm cov}_{\rm score}^{i} \leq {\rm cov}_{\rm min} \}.
\]

\paragraph{Merging Missing Information to Summary.}

We use perplexity ($\rm PPL$) to select the best location to insert a missing key sentence $k^i_{\rm missing} \in K_{\rm missing}$ into our summary~\cite{sharma2024textqualitybasedpruningefficient}:
\begin{eqnarray}
l^* = \operatorname*{argmin}_{l \in \rm locs} {\rm PPL}_{\rm LM}(k^i_{\rm missing}, {\rm summary},l ),
\end{eqnarray}
where ${\rm summary}$ is the summary obtained from Stage 2. We employ a greedy algorithm to dynamically insert the missing information. The complete algorithm will be provided in the appendix. 

\section{Evaluation}
We compare state-of-the-art approaches to medical summarization and improve the best-performing ones using \frameworkname{}. We demonstrate the improvements both through quantitative measurements and qualitative insights from a study conducted by domain experts.
\begin{figure*}[ht]
    \centering
    \includegraphics[width=1\linewidth]{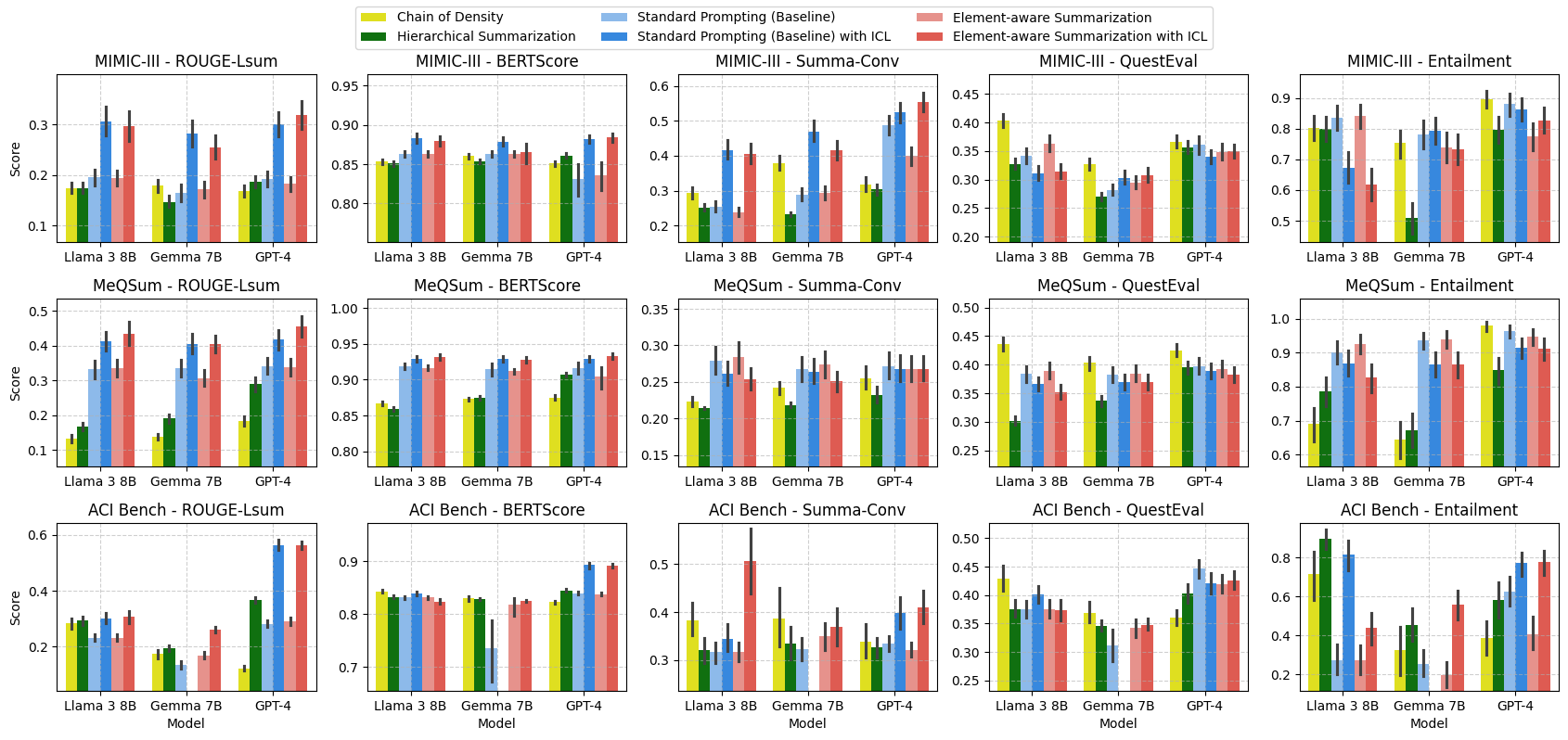}
    \caption{Benchmark of different Summarization Techniques across datasets on selected metrics. ROUGE-LSum and BertScore are reference-based metrics, while SummaC, QuestEval, and Entailment are used as reference-free metrics.}
    \label{fig:s1_benchmark}
\end{figure*}

\paragraph{Dataset and Tasks}

Figure \ref{fig:benchmark} describes the datasets, models and techniques chosen for our experimental setup. We make use of \textbf{three} biomedical datasets for summarization tasks: MIMIC III for Radiology Report Summarization \cite{johnson2016mimic}, MeQSum for Patient Question Summarization \cite{abacha2019summarization}, and ACI-Bench for doctor-patient dialogue summarization \cite{yim2023aci}. 
These provide a diverse range of biomedical summarization task settings, with varying document lengths, requirement for background knowledge as well as the need for domain-specific vocabulary and understanding. 

\paragraph{Evaluation Metrics}

 \citet{maynez-etal-2020-faithfulness} found that reference-based metrics by themselves do not align with human perception of faithfulness and factuality in abstractive summarization tasks and should be combined with reference-free metrics. We thus make use of \textbf{two reference-based metrics}, ROUGE-LSum \cite{lin2004rouge} and BERTScore \cite{zhang2019bertscore} which assess content overlap and semantic similarity with the reference summary, in combination with \textbf{three reference-free metrics}, SummaC \cite{laban2022summac}, QuestEval \cite{scialom2021questeval}, and Entailment Scores \cite{liu2024hit} which evaluate factual consistency, informativeness, and entailment relative to the source document for purposes of our evaluation. 

\paragraph{Experiment Setup}

\begin{table*}[ht]
\centering
\resizebox{\textwidth}{!}{
\input{table_v1_original}
}
\caption{Quantitative experiments showing the full performance and aggregated ranking of the \frameworkname{} pipeline on the three datasets. The base LLM and the method used are also included. The ranking column represents the aggregated ranks for all the metrics. The table contains two rankings---with and without entailment consideration---for an objective comparison. Numbers in \textcolor{red}{\underline{\smash{red + underlined}}} represent the best performance for the column, and those in \textcolor{red}{red} indicate the second-best performance. Numbers in \textcolor{blue}{\underline{blue + underlined}} represent the worst performance for the column, and those in \textcolor{blue}{blue} represent the second worst performance. \frameworkname{} and ICL methods perform well across most metrics, models, and datasets, with most of the relatively worse performances coming from existing summarization methods without the use of ICL.}
\label{tab:full_results}
\end{table*}

 We benchmark the performance of \textbf{four models}: LLaMA3 (8B) \cite{meta2024introducing}, Gemma (7B) \cite{team2024gemma}, Meditron (7B) \cite{chen2023meditron70b, epfmedtrn}, and GPT-4 \cite{achiam2023gpt}. 

In Stage 2, we use two NLI models for our experiments, DeBERTa v3 finetuned on NLI datasets \cite{he2021deberta, laurer2023deberta}, as well as biomedical finetuned PubMedBERT NLI model \cite{pubmedbert, lighteternal2023biomednlp}. To obtain the DSU's, sentence level decomposition is performed using PySBD \cite{sadvilkar2020pysbd} and atomic fact decomposition is obtained from the model used in Stage 1. 
Additionally, the modular setup of \frameworkname{} allows us to compare the confabulation detection of dedicated NLI models with LLM based techniques such as Self-Reflection~\citep{ji2023towards} as an ablation study.

For Stage 3, we use all-MiniLM-L6-v2 \cite{wang2020minilm} as the encoder for key information extraction and missing information detection, following the Sentence-BERT \cite{reimers-2019-sentence-bert} framework. Additionally, we set $\rm cov_{\rm min}$ to 0.4; this parameter can be tuned based on the specific coverage metrics or models employed. When merging missing information, we use GPT-2 \cite{radford2019language} to calculate the perplexity and rearrange the added sentences to improve fluency and coherence~\cite{sharma2024textqualitybasedpruningefficient}.

Formally, let \( S_i \) represent the summary generated by the \( i \)-th summarization technique. The quality of this summary is evaluated using a set of metrics \( M_j \), where \( j \) denotes the specific metric used (e.g., ROUGE-L, BERTScore, SummaC). The score for a given metric \( M_j \) applied to a summary \( S_i \) is denoted as \( M_j(S_i) \). The final evaluation score for a summary \( S_i \) generated by a specific method is determined by aggregating its rank across all metrics:
\begin{eqnarray}
\text{Rank}_i = \sum_{j=1}^{n} \text{Rank}(M_j(S_i)),
\end{eqnarray}
where \(\text{Rank}(M_j(S_i))\) is the rank of the method based on metric \( M_j \), and \( n \) is the total number of metrics used. The method with the lowest \(\text{Rank}_i\) is considered the most effective. 
Since NLI is used directly in confabulation detection in Stage 2, we provide two separate rankings for a more objective comparison: one considering entailment and one without. 

\section{Results and Discussion}

\subsection{Summarization Techniques Benchmark} 
\label{sec:results_discussion}

Figure \ref{fig:s1_benchmark} presents the benchmark results of different summarization techniques. Table \ref{tab:full_results} presents the full set of results for the benchmark, as well as results of \frameworkname{} based experiments. The benchmark results are structured into three key areas: the performance of summarization methods, the influence of datasets, and the comparative evaluation of models. 

\paragraph{Methods.} The Standard Prompting method was selected as the baseline. 
We first compare the different summarization techniques in zero-shot settings without task adaptation.
Chain of Density particularly improves the performance of the QuestEval metric, which aims to measure the factual information retention between input documents and summaries.
This can be attributed to its nature of creating the most information-dense summaries, albeit at the cost of readability and conciseness.
The Hierarchical method demonstrates noticeable performance in summarizing longer documents, effectively mitigating the ``lost-in-the-middle'' effect \cite{ravaut2023context}.
This is evidenced by consistent improvements across all models when employing hierarchical summarization, particularly for lengthy inputs like those in ACI Bench.
The approach enhances faithfulness to the input document of summaries by decomposing documents into manageable blocks.
While these methods excel in specific areas, Element Aware Summarization outperforms them by leveraging model reasoning to extract the most relevant information, summarize it effectively, and achieve the best rank across all models.

Task adaptation using ICL consistently improves both reference-based and reference-free metrics across all datasets. Our findings align with~\citet{van2024adapted}, confirming GPT-4 Standard Prompting with ICL as the previous SOTA for medical summarization.
However, our benchmark reveals that Element Aware Summarization with ICL-based task adaptation surpasses this previously established SOTA. 
This demonstrates that task adaptation complements model reasoning techniques in enhancing summary quality.

\paragraph{Datasets.}
Figure \ref{fig:s1_benchmark} suggests that for MIMIC-III, models perform worse on phrase overlap metrics such as ROUGE-LSum while maintaining relatively high scores in reference-free and reference-based semantic metrics such as BERTScore and SummaC, indicating that the models tend to paraphrase or compress the information in the input document while staying consistent and faithful to the inputs for MIMIC-III. 

MeQSum contains the shortest input documents and involves summarizing patient questions, which requires less background knowledge but a clear understanding of the query.
The models perform the best on MeQSum across most metrics, particularly in reference-free metrics like QuestEval and Entailment, reflecting the models' ability to handle content low on domain-specific jargon.
Notably, the shorter and less technical nature of MeQSum allows smaller models like Gemma 7B and Llama 3 8B to perform competitively with GPT-4, as the task requires a clear understanding of short queries rather than extensive domain knowledge or information extraction capabilities.

Due to its conversational nature and length, ACI Bench requires the summarization of long context documents that might include more redundant or less structured information.
Task adaptation using ICL particularly helps in this dataset, where giving the models examples of the kind of information to focus on in the input significantly improves performance.
The models suffer in all reference-free metrics, indicating that the ACI bench is particularly difficult for the models in extracting and relaying key information in the summary while staying faithful to the source document.

\paragraph{Models.} As shown in Table \ref{tab:full_results}, Meditron 7B exhibits the lowest performance across most metrics and datasets when using Standard Prompting.
Due to its limited instruction-following capabilities, we only consider Meditron for Standard Prompting.
Gemma 7B shows weak performance in metrics like RougeLSum and SummaC.
It particularly struggles with Entailment across all datasets, showing poor ability to maintain logical consistency and faithfulness in summaries.
Llama 3 8B performs the best among the open-source models, often showing competitive performance to GPT-4 for the given tasks. Llama 3 8B also benefits more from ICL than Gemma, which highlights its strong ability to adapt to tasks. Consistent with the findings of \citet{van2024adapted}, we find that unsurprisingly, GPT-4 performs best across all summarization tasks. 
Overall, we find that GPT-4 performs best, with Llama3 8B being the best-performing open-source model.
Based on these findings, we select the two models: Llama 3 8B and GPT-4, and two methods: the previously established SOTA of Standard Prompting as well as the best-performing technique based on our experiments, Element Aware Summarization, for the next stage of experiments using \frameworkname{}.    

\subsection{Analysis of \frameworkname{} Results} 

Table~\ref{tab:full_results} demonstrates that \textbf{\frameworkname{} consistently outperforms the above-mentioned benchmark results}, with seven out of the top ten ranked methods utilizing \frameworkname{}.
We especially see significant improvement in reference-free metrics that assess the factual consistency and completeness of the summaries, such as SummaC, QuestEval, and Entailment, while being competitive or improving performance in reference-based metrics. This indicates that \frameworkname{} improves faithfulness and informativeness of the summaries, while staying grounded to the input document.
\frameworkname{}'s impact is most pronounced when combined with Element Aware Summarization and ICL, suggesting that it can be used in combination with methods leveraging model reasoning as well as task adaptation techniques to produce summaries that utilize the key benefits of all the methods. 

For all datasets, \frameworkname{} helps improve ROUGE-LSum, particularly with Llama 3 8B.
\frameworkname{} also maintains a high BertScore across datasets, particularly with GPT-4.
This suggests that the additional stages of confabulated information removal and missing information addition preserve and even enhance semantic similarity between generated and reference summaries by focusing on error correction and gap-filling.
Notably, SummaC and Entailment scores significantly improve for all models when using \frameworkname{}. 
These metrics directly benefit from the confabulation detection and removal stage, as they ensure that the final summary is factually consistent and faithful to the source information.
QuestEval scores show marked improvements as well.
The missing information addition stage (Stage 3) proves particularly beneficial, ensuring comprehensive coverage of key aspects of the input document.
Lastly, we point out that  \frameworkname{} significantly improves summarization quality for smaller models.
For instance, Llama3 8B with \frameworkname{} and ICL outperforms GPT-4's Standard Prompting baseline and remains competitive with GPT-4 across all metrics, despite starting from a significantly lower baseline performance.
\paragraph{Ablation Studies.}
We conducted ablation studies by removing individual stages and comparing performance against the complete framework.
Results show that Stages 2 and 3 complement each other, with net gains across both reference-based and reference-free metrics, leading to more comprehensive and faithful summaries.
Additional ablations explored different NLI models \cite{pubmedbert, laurer2023deberta} and LLM-based hallucination removal methods like self-reflection \cite{ji2023towards}.
Stage 2 using a DeBERTa-based finetuned NLI model \cite{laurer2023deberta} performed best across datasets and models.
Full ablation results are provided in the technical appendix.

\subsection{Clinician Evaluation}
We perform a human evaluation by two orthopaedic surgeons for the radiology report summarization task, who are provided with related summaries generated using the previous SOTA (Standard Prompting ICL + GPT-4), and our best performing method (Element Aware + ICL \frameworkname{} + GPT-4). 
Doctors performed pairwise selections based on overall summary quality and annotated \textbf{difficult cases} with confabulations or missing key information, without knowing the methods which generated the summaries.
Our results show that when there were no confabulations or missing information, doctors showed equal preference between the previous SOTA and \frameworkname{}. However, in \textbf{difficult cases} involving confabulations or missing key information, doctors preferred \frameworkname{} 46\% of the time, citing its effectiveness in resolving issues, compared to only 8\% for the previous SOTA. Both summaries were considered inadequate 23\% of the time, acceptable 15\%, and undecidable in 8\% of cases. 
This preference underscores the critical importance of \frameworkname{} in minimizing errors in medical summaries, as the impact of resolving confabulations or missing information far outweighs the benefit of matching previous methods in straightforward cases when considering patient care.
Full clinical study details are provided in the appendix.

\section{Conclusion}
We introduce \frameworkname{}, a novel framework for accurate and informative medical summarization. We conduct a comprehensive benchmark and integrate the findings with \frameworkname{} to surpass recent SOTA on medical summarization. We achieve a significant 11.8\% improvement in reference-free metrics which focus on faithfulness and informativeness without sacrificing reference-based performance. The human evaluation shows doctors preferred \frameworkname{} six times more than the previous SOTA 
in the presence of confabulations or the absence of key information.
Our approach sets a new standard for faithful and informative medical summarization.
\section*{Acknowledgements}
Viktor is supported by the National Research Foundation, Prime Minister’s Office, Singapore under its Campus for Research Excellence and Technological Enterprise (CREATE) programme.

\bibliography{aaai25,references}

\clearpage
\appendix

\title{Appendix}

\section{Algorithms}

\subsection{Algorithm 1}

\input{algo_1_key_info_extraction} 

\subsection{Algorithm 2}

\begin{table*}[!t]
\centering
\resizebox{\textwidth}{!}{
\input{ablations_table}
}
\caption{Quantitative experiments showing the full performance and aggregated ranking of ablation studies on the \frameworkname{} pipeline on the three datasets. The base LLM and the method used are also included. The ranking column represents the aggregated ranks for all the metrics. The table contains two rankings---with and without entailment consideration---for an objective comparison. Numbers in \textcolor{red}{\underline{\smash{red + underlined}}} represent the best performance for the column, and those in \textcolor{red}{red} indicate the second-best performance. Numbers in \textcolor{blue}{\underline{blue + underlined}} represent the worst performance for the column, and those in \textcolor{blue}{blue} represent the second worst performance. The best performing combination on all the datasets was used for the main set of experiments provided in the paper. We see here that Element Aware + ICL + Stage 3 using DeBERTA NLI Model + Stage 3 is the best performing model based on the the ablation studies. }
\label{tab:ablation_results}
\end{table*}

\input{algo_2_merge_missing_sentences} 

\section{Clinical Evaluation}
Our evaluators include two orthopaedic surgeons, who are provided with related summaries generated using 2 methods:

\begin{enumerate}
    \item Standard Prompting ICL + GPT-4 (Previous SOTA) \label{item:standard}
    \item Element Aware + ICL \frameworkname{} + GPT-4 (overall best performing method) \label{item:framework}
\end{enumerate}

We specifically selected summaries relevant to the clinicians in order to fully utilize their expertise during evaluation. Thus, from the subset of MIMIC-III \cite{johnson2016mimic} which was used for the experiments, we selected a subset of 60 samples filtered using the fillowing keywords: Arthritis, Bone, Clavicle, Deformity, Dislocation, Femur, Fibula, Fracture, Humerus, Intervertebral Disc, Joint, Ligament, Malunion, Non-union, Osteophyte, Patella, Radius, Sacrum, Scapula, Scoliosis, Sondylolisthesis, Spondylosis, Spine, Spur, Tibia, Ulna, Union. 

The clinicians were asked to select the preferred summary in pairwise fashion. This meant that for each input document, they would be provided 2 summaries: A summary generated by method \ref{item:standard} and another by method \ref{item:framework}. The clinicians would not be aware of the model or technique which generated the summaries, in order to avoid any bias. The clinicans were asked to evaluate both the summaries according to the following criteria:

\begin{enumerate}
    \item Which summary do they prefer between the two summaries?
    \item Is there any information which should be removed from either of the summaries?
    \item Is there any key information missing from either of the summaries?
\end{enumerate}

This allowed us to measure the general quality of the summary, confabulations in generated summaries, missing key information in the generated summaries. We only selected cases where the clinicians had a consensus on the preference between both the summaries. We classify \textbf{difficult} cases as cases where either of the summaries contains confabulations or missing information according to any of the clinicians based on their annotations. If not, the summaries are considered straightforward.

\section{Hyperparameter Search}
For Stage 2 and Stage 3 of \frameworkname{}, we perform a qualitative analysis of threshold values. For Stage 2, we perform grid search for Entailment ($T_e$), Contradiction ($T_c$), Atomic Fact Entailment ($T_a$), while for stage 3, we perform grid search for Number of Key Sentences to Extract (${\rm top_M}$) and Minimum Coverage Score (${\rm cov}_{\rm min}$) thresholds. This analyis is performed on a subset of all the datasets used for evaluation, but sample non-intersecting, separate data points from each dataset for fixing thresholds so as to not overfit on the test sample. 40 such data points from each dataset are used for fixing the thresholds. We used summaries generated by GPT-4 + ICL + \frameworkname{} for the threshold selection process.\\ For Stage 2, in order to optimize the process of selecting the optimal thresholds, we first start with $T_e$ and $T_c$ fixed to their most extreme values to maximise the condition for "uncertain" DSU's to be split into atomic fact. In this setting, we then find the optimal $T_a$ which can be reasonably used without overzealous removal or retention of information presented in atomic facts. Next, We fix $T_e$ and $T_a$ and find the optimal values for $T_c$, and finally use the same process to find the optimal values for $T_e$. The final thresholds used for Stage 2 are provided below:
\begin{eqnarray}
T_e &=& 0.9,\\
T_c &=& 0.8,\\
T_a &=& 0.5.
\end{eqnarray}

Similar results were found using both the NLI models used in the ablation studies, which can be attributed to almost bimodal distribution of entailed and contradicting facts in the generated summaries. \\

For Stage 3, in order to optimize the process of selecting the optimal thresholds, we first keep ${\rm cov}_{\rm min}$ fixed to higher bound to maximise the chance that $k_{\rm doc}^i$ is selected as missing information and find the optimal ${\rm top_M}$. Next, we fix ${\rm top_M}$ to find optimal values for ${\rm cov}_{\rm min}$. The final tuned thresholds used for Stage 3 are provided below:
\begin{eqnarray}
{\rm top_M} &=& 2,\\
{\rm cov}_{\rm min} &=& 0.4.
\end{eqnarray}

\section{Prompts}

Table \ref{tab:prompts} provides the dataset specific prompts used for the experiments. These prompts were used as-is for the Standard Prompting experiments. For other methods, these prompts were combined with method specific instructions and logic, which can be found in the code.

\begin{table}
\caption{Summary of prompts}
\label{tab:prompts}
\begin{tabularx}{\linewidth}{|>{\RaggedRight}X|>{\RaggedRight\arraybackslash}X|}
\hline
\textbf{Key} & \textbf{Description} \\
\hline
\texttt{MIMIC-III} & Summarize the radiology report findings into an impression in 35 words or less \\
\hline
\texttt{MeQSum} & Summarize the patient health query into one question of 15 words or less \\
\hline
\texttt{ACI-Bench} & Summarize the patient/doctor dialogue into an assessment and plan \\
\hline
\end{tabularx}
\end{table}

\begin{samepage}
\section{Ablation Study}

Table \ref{tab:ablation_results} presents the full ablation study performed. We start from the best performing summarisation method from the benchmark - Element Aware Summarisation, and first add each stage of \frameworkname{} separately. Finally, we test on separate configurations of confabulation removal on the end-to-end \frameworkname{} pipeline. 
We conduct first conduct an ablation study using Standard Prompting + ICL which was established as the previous SOTA, and comparing the impact of LLM based techniques such as self-reflection with our proposed Stage-2 using DeBERTa. We find that for standard prompting, self-reflection performs better. 

We then conduct a full ablation study using our best performing technique (Element Aware Summarisation). In this ablation, we begin with the base Element Aware method, and incrementally add different stages of \frameworkname{} to the generated summary. Thus, in the next step, we implement Stage 2 with DeBERTa as well as Self-reflection, followed by task adaptation using ICL. Since the results suggest that NLI based Stage 2 performs better than self-reflection based stage 2, we use DeBERTa based stage 2 and evaluate the impact of Element Aware + ICL with stage 2 and stage 3 separately. 

Finally, we implement ablations on the final framework, where we again implement 2 different NLI models as well as self-reflection and evaluate the impact on the final \frameworkname{} results. The ablation results suggest that the final \frameworkname{} gives the most balanced results, with all 5 ablations using the full framework achieving 5 out of the top 6 ranks. Additionally, using Stage 2 and Stage 3 together complements performance, as seen from the fact that there is an improvement in both reference based as well as reference free metrics in Element Aware + ICL + Stage 2 (DeBERTa) + Stage 3 as compared to Element Aware + ICL + Stage 2 (DeBERTa) and Element Aware + ICL + Stage 3. 

We also confirm that Element Aware outperforms Standard Prompting (Previous SOTA) for both NLI based as well as self-reflection based methods for Stage 2, further reinforcing the benchmark results. 

The results show that the best rank is obtained by \frameworkname{} using DeBERTa as the NLI module for confabulation detection in Stage 2 achieves the best performance. The experiments with Stage 3 are ranked high since they directly optimize for the reference based metrics, which might not always necessarily lead to better abstractive summaries but do improve the reference based metrics. 

Considering a holistic improvement across both reference based and reference free metrics, the end-to-end pipeline for \frameworkname{} still performs the best. 

\end{samepage}

\end{document}

%% file: table_v1_original.tex
\begin{tabular}{p{3.3cm}llllllllllllllllcc}
\toprule
 & \multicolumn{1}{r}{Dataset:} & \multicolumn{5}{c}{MIMIC-III} & \multicolumn{5}{c}{ACI Bench} & \multicolumn{5}{c}{MeQSum} & \multicolumn{2}{c}{Ranking} \\
 \cmidrule(lr){3-7}\cmidrule(lr){8-12}\cmidrule(r){13-17}\cmidrule(r){18-19}
Method + Model & \multicolumn{1}{r}{Metric:} & R-Ls & B.S. & S-C & Q.E. & Ent. & R-Ls & B.S. & S-C & Q.E. & Ent. & R-Ls & B.S. & S-C & Q.E. & Ent. & w/ Ent. & w/o Ent.  \\
\midrule
\multirow[c]{4}{*}{\parbox{2.7cm}{\footnotesize Standard Prompting (Baseline)}} & Meditron 7B & \textbf{\underline{\textcolor{blue}{0.09}}} & \textbf{\underline{\textcolor{blue}{0.81}}} & 0.47 & 0.33 & \textbf{\textcolor{blue}{0.61}} & \textbf{\underline{\textcolor{blue}{0.10}}} & \textbf{\underline{\textcolor{blue}{0.65}}} & 0.51 & \textbf{\underline{\textcolor{blue}{0.25}}} & 0.30 & \textbf{\underline{\textcolor{blue}{0.04}}} & \textbf{\underline{\textcolor{blue}{0.80}}} & 0.27 & \textbf{\underline{\textcolor{blue}{0.26}}} & \textbf{\underline{\textcolor{blue}{0.60}}} & \textbf{\textcolor{blue}{24}} & 22 \\
 & Gemma 7B & 0.16 & 0.86 & 0.29 & \textbf{\textcolor{blue}{0.28}} & 0.78 & 0.13 & 0.74 & 0.32 & 0.31 & \textbf{\textcolor{blue}{0.25}} & 0.34 & 0.92 & 0.27 & 0.38 & 0.94 & 23 & 23 \\
 & Llama 3 8B & 0.20 & 0.86 & 0.25 & 0.34 & 0.83 & 0.23 & 0.83 & \textbf{\underline{\textcolor{blue}{0.32}}} & 0.38 & 0.27 & 0.33 & 0.92 & 0.28 & 0.38 & 0.90 & 16 & 15 \\
 & GPT-4 & 0.19 & \textbf{\textcolor{blue}{0.83}} & 0.49 & 0.36 & 0.88 & 0.28 & 0.84 & 0.33 & 0.45 & 0.63 & 0.34 & 0.92 & 0.27 & 0.40 & 0.96 & 8 & 10 \\
\cline{1-19}
\multirow[c]{3}{*}{\parbox{3cm}{\footnotesize Chain of Density}} & Gemma 7B & 0.18 & 0.86 & 0.38 & 0.33 & 0.75 & 0.17 & 0.83 & 0.39 & 0.37 & 0.32 & 0.14 & 0.87 & 0.24 & 0.40 & \textbf{\textcolor{blue}{0.65}} & 21 & 20 \\
 & Llama 3 8B & 0.17 & 0.85 & 0.29 & \textbf{\underline{\textcolor{red}{0.40}}} & 0.80 & 0.28 & 0.84 & 0.38 & 0.43 & 0.71 & \textbf{\textcolor{blue}{0.13}} & 0.87 & 0.22 & \textbf{\underline{\textcolor{red}{0.44}}} & 0.69 & 15 & 16 \\
 & GPT-4 & 0.17 & 0.85 & 0.32 & 0.37 & \textbf{\underline{\textcolor{red}{0.90}}} & \textbf{\textcolor{blue}{0.12}} & 0.82 & 0.34 & 0.36 & 0.38 & 0.18 & 0.88 & 0.26 & \textbf{\textcolor{red}{0.42}} & \textbf{\underline{\textcolor{red}{0.98}}} & 19 & 21 \\
\cline{1-19}
\multirow[c]{3}{*}{\footnotesize Hierarchical} & Gemma 7B & \textbf{\textcolor{blue}{0.15}} & 0.85 & \textbf{\underline{\textcolor{blue}{0.23}}} & \textbf{\underline{\textcolor{blue}{0.27}}} & \textbf{\underline{\textcolor{blue}{0.51}}} & 0.19 & 0.83 & 0.33 & 0.35 & 0.45 & 0.19 & 0.87 & \textbf{\textcolor{blue}{0.22}} & 0.34 & 0.67 & \textbf{\underline{\textcolor{blue}{25}}} & \textbf{\underline{\textcolor{blue}{25}}} \\
 & Llama 3 8B & 0.17 & 0.85 & 0.25 & 0.33 & 0.80 & 0.29 & 0.83 & \textbf{\textcolor{blue}{0.32}} & 0.38 & 0.90 & 0.17 & \textbf{\textcolor{blue}{0.86}} & \textbf{\underline{\textcolor{blue}{0.21}}} & \textbf{\textcolor{blue}{0.30}} & 0.79 & 21 & \textbf{\textcolor{blue}{24}} \\
 & GPT-4 & 0.19 & 0.86 & 0.30 & 0.36 & 0.80 & 0.37 & 0.84 & 0.33 & 0.40 & 0.58 & 0.29 & 0.91 & 0.23 & 0.40 & 0.85 & 13 & 13 \\
\cline{1-19}
\multirow[c]{3}{*}{\footnotesize Element Aware} & Gemma 7B & 0.17 & 0.86 & 0.29 & 0.29 & 0.74 & 0.17 & 0.82 & 0.35 & 0.34 & \textbf{\underline{\textcolor{blue}{0.20}}} & 0.31 & 0.91 & 0.27 & 0.38 & 0.94 & 20 & 19 \\
 & Llama 3 8B & 0.19 & 0.86 & \textbf{\textcolor{blue}{0.24}} & 0.36 & 0.84 & 0.23 & 0.83 & \textbf{\underline{\textcolor{blue}{0.32}}} & 0.38 & 0.27 & 0.33 & 0.92 & 0.28 & 0.39 & 0.93 & 16 & 16 \\
 & GPT-4 & 0.18 & 0.84 & 0.40 & 0.35 & 0.78 & 0.29 & 0.84 & 0.32 & 0.42 & 0.41 & 0.34 & 0.90 & 0.27 & 0.39 & 0.95 & 14 & 14 \\
\cline{1-19}
\multirow[c]{2}{*}{\parbox{2.6cm}{\footnotesize Element Aware + \frameworkname{} (\textbf{Ours})}} & Llama 3 8B & 0.19 & 0.86 & 0.44 & 0.39 & 0.86 & 0.15 & 0.81 & 0.53 & 0.44 & 0.67 & 0.31 & 0.91 & 0.35 & 0.41 & 0.93 & 10 & 11 \\
 & GPT-4 & 0.19 & 0.86 & 0.53 & 0.39 & 0.89 & 0.15 & 0.82 & 0.51 & \textbf{\underline{\textcolor{red}{0.47}}} & 0.86 & 0.33 & 0.91 & 0.33 & 0.41 & \textbf{\textcolor{red}{0.97}} & 6 & 7 \\ \midrule \midrule
\multirow[c]{3}{*}{\parbox{2.65cm}{\footnotesize Standard Prompting (Baseline) + ICL}} & Gemma 7B & 0.28 & 0.88 & 0.47 & 0.30 & 0.79 & 0.17 & \textbf{\textcolor{blue}{0.72}} & 0.33 & \textbf{\textcolor{blue}{0.30}} & 0.40 & 0.41 & 0.93 & 0.26 & 0.37 & 0.87 & 18 & 18 \\
 & Llama 3 8B & \textbf{\textcolor{red}{0.31}} & \textbf{\textcolor{red}{0.88}} & 0.42 & 0.31 & 0.67 & 0.30 & 0.84 & 0.34 & 0.40 & 0.82 & 0.41 & 0.93 & 0.26 & 0.37 & 0.87 & 9 & 9 \\
 & GPT-4 & 0.30 & 0.88 & 0.52 & 0.34 & 0.86 & \textbf{\underline{\textcolor{red}{0.56}}} & \textbf{\underline{\textcolor{red}{0.89}}} & 0.40 & 0.42 & 0.77 & 0.42 & 0.93 & 0.27 & 0.39 & 0.91 & 4 & 4 \\
\cline{1-19}
\multirow[c]{3}{*}{\footnotesize Element Aware + ICL} & Gemma 7B & 0.25 & 0.87 & 0.41 & 0.31 & 0.73 & 0.26 & 0.83 & 0.37 & 0.35 & 0.56 & 0.40 & 0.93 & 0.25 & 0.37 & 0.86 & 12 & 12 \\
 & Llama 3 8B & 0.30 & 0.88 & 0.41 & 0.31 & 0.62 & 0.31 & 0.82 & 0.51 & 0.37 & 0.44 & \textbf{\textcolor{red}{0.43}} & \textbf{\textcolor{red}{0.93}} & 0.25 & 0.35 & 0.83 & 11 & 8 \\
 & GPT-4 & \textbf{\underline{\textcolor{red}{0.32}}} & \textbf{\underline{\textcolor{red}{0.88}}} & 0.55 & 0.35 & 0.83 & \textbf{\textcolor{red}{0.56}} & \textbf{\textcolor{red}{0.89}} & 0.41 & 0.43 & 0.78 & \textbf{\underline{\textcolor{red}{0.46}}} & \textbf{\underline{\textcolor{red}{0.93}}} & 0.27 & 0.38 & 0.91 & 3 & 3 \\
\cline{1-19}
\multirow[c]{2}{*}{\parbox{3.5cm}{\footnotesize Standard Prompting + ICL + \frameworkname{} (\textbf{Ours})}} & Llama 3 8B & 0.25 & 0.87 & 0.56 & 0.37 & 0.74 & 0.16 & 0.82 & \textbf{\textcolor{red}{0.57}} & 0.45 & \textbf{\underline{\textcolor{red}{0.96}}} & 0.42 & 0.92 & \textbf{\underline{\textcolor{red}{0.36}}} & 0.38 & 0.84 & 5 & 5 \\
 & GPT-4 & 0.25 & 0.87 & \textbf{\textcolor{red}{0.61}} & 0.39 & \textbf{\textcolor{red}{0.89}} & 0.44 & 0.86 & 0.48 & 0.45 & 0.91 & 0.38 & 0.92 & 0.35 & 0.41 & 0.89 & \textbf{\textcolor{red}{2}} & \textbf{\textcolor{red}{2}} \\
\cline{1-19}
\multirow[c]{2}{*}{\parbox{3.5cm}{\footnotesize Element Aware + ICL + \frameworkname{} (\textbf{Ours})}} & Llama 3 8B & 0.25 & 0.87 & 0.58 & 0.37 & 0.71 & 0.17 & 0.81 & \textbf{\underline{\textcolor{red}{0.63}}} & 0.42 & 0.77 & 0.39 & 0.92 & \textbf{\textcolor{red}{0.35}} & 0.37 & 0.81 & 7 & 6 \\
 & GPT-4 & 0.25 & 0.87 & \textbf{\underline{\textcolor{red}{0.64}}} & \textbf{\textcolor{red}{0.39}} & 0.88 & 0.42 & 0.86 & 0.49 & \textbf{\textcolor{red}{0.45}} & \textbf{\textcolor{red}{0.92}} & 0.41 & 0.92 & 0.34 & 0.40 & 0.89 & \textbf{\underline{\textcolor{red}{1}}} & \textbf{\underline{\textcolor{red}{1}}} \\
\cline{1-19}
\bottomrule
\end{tabular}

%% file: algo_1_key_info_extraction.tex
In Algorithm \ref{alg:1}, ${\rm top}_K$ represents the threshold for the number of key sentences or key phrases to extract. $K$ is a list of key sentences extracted from input document or key phrases extracted from Stage 2 summary, and ${\rm len}(K)$ represents the number of extracted key sentences or key phrases in the $K$.

\begin{algorithm}[H]
\caption{Key Information Extraction}
\label{alg:1}
\begin{algorithmic}[1]
  \STATE ${\rm input} \gets  \text{document or Stage 2 summary}$
  \STATE $K \gets \{\}$
  \IF {$\rm input = \text{document}$} 
    \STATE ${\rm candidates} \gets \text{each sentence in document}$ 
  \ELSE
    \STATE ${\rm candidates} \gets \text{each phrase in summary}$ 
  \ENDIF
  \WHILE{${\rm len}(K) \leq {\rm top}_K$}
    \STATE \(
      {\rm candidate}^* := \operatorname*{argmax}_{x \in {\rm candidates}} {\rm MMR(input, } K,  x)
    \)
    \STATE $ K \gets K + \rm candidate^*$
  \ENDWHILE
  \RETURN $K$
\end{algorithmic}
\end{algorithm}

%% file: ablations_table.tex
\begin{tabular}{lllllllllllllllllll}
\toprule
 & \textbf{Dataset} & \multicolumn{5}{r}{\textbf{MIMIC-III}} & \multicolumn{5}{r}{\textbf{ACI Bench}} & \multicolumn{5}{r}{\textbf{MeQSum}} & & \\
 \textbf{Method} & \textbf{Metric} & R-Ls & B.S. & S-C & Q.E. & Ent. & R-Ls & B.S. & S-C & Q.E. & Ent. & R-Ls & B.S. & S-C & Q.E. & Ent. &  \textbf{Rank 1} & \textbf{Rank 2}  \\
 & \textbf{Model} &  &  &  &  &  &  &  &  &  &  &  &  &  &  &  &  &  \\
\midrule
\multirow[t]{2}{*}{Standard Prompting + ICL + Stage 2 (Deberta) + Stage 3} & Llama 3 8B & 0.25 & 0.87 & 0.56 & 0.37 & 0.74 & 0.16 & 0.82 & 0.57 & 0.45 & \textbf{\underline{\textcolor{red}{0.96}}} & 0.42 & 0.92 & 0.36 & 0.38 & 0.84 & 8.00 & 7.00 \\
 & GPT-4 & 0.25 & 0.87 & 0.61 & 0.39 & \textbf{\underline{\textcolor{red}{0.89}}} & 0.44 & 0.86 & 0.48 & 0.45 & 0.91 & 0.38 & 0.92 & 0.35 & 0.41 & 0.89 & 3.00 & 5.00 \\
\cmidrule{1-19}
\multirow[t]{2}{*}{Standard Prompting + ICL + Stage 2 (Reflection) + Stage 3} & Llama 3 8B & 0.25 & 0.87 & 0.59 & 0.38 & 0.70 & 0.27 & 0.83 & 0.44 & 0.42 & 0.75 & 0.33 & 0.90 & 0.38 & 0.38 & \textbf{\textcolor{blue}{0.68}} & 13.00 & 11.00 \\
 & GPT-4 & 0.26 & 0.87 & 0.62 & 0.38 & 0.88 & \textbf{\underline{\textcolor{red}{0.91}}} & \textbf{\underline{\textcolor{red}{0.93}}} & 0.40 & 0.41 & 0.81 & 0.38 & 0.92 & 0.35 & 0.40 & 0.89 & 6.00 & 6.00 \\
\cmidrule{1-19}
\multirow[t]{3}{*}{Element Aware} & Llama 3 8B & 0.19 & 0.86 & \textbf{\underline{\textcolor{blue}{0.24}}} & 0.36 & 0.84 & 0.23 & 0.83 & \textbf{\underline{\textcolor{blue}{0.32}}} & 0.38 & \textbf{\textcolor{blue}{0.27}} & 0.33 & 0.92 & 0.28 & 0.39 & 0.93 & \textbf{\textcolor{blue}{22.00}} & 21.00 \\
 & GPT-4 & \textbf{\textcolor{blue}{0.18}} & \textbf{\underline{\textcolor{blue}{0.84}}} & 0.40 & 0.35 & 0.78 & 0.29 & 0.84 & \textbf{\textcolor{blue}{0.32}} & 0.42 & 0.41 & 0.34 & 0.90 & 0.27 & 0.39 & 0.95 & \textbf{\textcolor{blue}{22.00}} & 22.00 \\
 & Gemma 7B & \textbf{\underline{\textcolor{blue}{0.17}}} & 0.86 & 0.29 & \textbf{\underline{\textcolor{blue}{0.29}}} & 0.74 & 0.17 & 0.82 & 0.35 & \textbf{\underline{\textcolor{blue}{0.34}}} & \textbf{\underline{\textcolor{blue}{0.20}}} & \textbf{\textcolor{blue}{0.31}} & 0.91 & 0.27 & 0.38 & 0.94 & \textbf{\underline{\textcolor{blue}{24.00}}} & \textbf{\underline{\textcolor{blue}{24.00}}} \\
\cmidrule{1-19}
\multirow[t]{2}{*}{Element Aware + Stage 2 (Deberta) + Stage 3} & Llama 3 8B & 0.19 & 0.86 & 0.44 & 0.39 & 0.86 & \textbf{\textcolor{blue}{0.15}} & 0.81 & 0.53 & 0.44 & 0.67 & 0.31 & 0.91 & 0.35 & 0.41 & 0.93 & 16.00 & 17.00 \\
 & GPT-4 & 0.19 & \textbf{\textcolor{blue}{0.86}} & 0.53 & 0.39 & \textbf{\textcolor{red}{0.89}} & 0.15 & 0.82 & 0.51 & \textbf{\underline{\textcolor{red}{0.47}}} & 0.86 & 0.33 & 0.91 & 0.33 & \textbf{\underline{\textcolor{red}{0.41}}} & \textbf{\underline{\textcolor{red}{0.97}}} & 10.00 & 14.00 \\
\cmidrule{1-19}
\multirow[t]{2}{*}{Element Aware + Stage 2 (Reflection) + Stage 3} & Llama 3 8B & 0.19 & 0.86 & 0.48 & \textbf{\underline{\textcolor{red}{0.40}}} & 0.86 & 0.21 & 0.82 & 0.47 & 0.43 & 0.38 & \textbf{\underline{\textcolor{blue}{0.28}}} & \textbf{\underline{\textcolor{blue}{0.90}}} & \textbf{\underline{\textcolor{red}{0.40}}} & 0.40 & 0.76 & 20.00 & 18.00 \\
 & GPT-4 & 0.19 & 0.86 & 0.52 & 0.39 & 0.83 & 0.30 & 0.83 & 0.37 & 0.45 & 0.65 & 0.32 & 0.91 & 0.32 & \textbf{\textcolor{red}{0.41}} & \textbf{\textcolor{red}{0.96}} & 11.00 & 13.00 \\
\cmidrule{1-19}
\multirow[t]{3}{*}{Element Aware + ICL} & Llama 3 8B & 0.30 & \textbf{\textcolor{red}{0.88}} & 0.41 & 0.31 & \textbf{\textcolor{blue}{0.62}} & 0.31 & 0.82 & 0.51 & 0.37 & 0.44 & 0.43 & 0.93 & \textbf{\textcolor{blue}{0.25}} & \textbf{\underline{\textcolor{blue}{0.35}}} & 0.83 & 18.00 & 16.00 \\
 & GPT-4 & \textbf{\underline{\textcolor{red}{0.32}}} & \textbf{\underline{\textcolor{red}{0.88}}} & 0.55 & 0.35 & 0.83 & 0.56 & 0.89 & 0.41 & 0.43 & 0.78 & \textbf{\underline{\textcolor{red}{0.46}}} & \textbf{\underline{\textcolor{red}{0.93}}} & 0.27 & 0.38 & 0.91 & 9.00 & 9.00 \\
 & Gemma 7B & 0.25 & 0.87 & 0.41 & \textbf{\textcolor{blue}{0.31}} & 0.73 & 0.26 & 0.83 & 0.37 & \textbf{\textcolor{blue}{0.35}} & 0.56 & 0.40 & 0.93 & \textbf{\underline{\textcolor{blue}{0.25}}} & \textbf{\textcolor{blue}{0.37}} & 0.86 & 17.00 & 19.00 \\
\cmidrule{1-19}
\multirow[t]{2}{*}{Element Aware + ICL + Stage 2 (Deberta)} & Llama 3 8B & 0.19 & 0.86 & \textbf{\textcolor{blue}{0.24}} & 0.36 & 0.87 & \textbf{\underline{\textcolor{blue}{0.12}}} & 0.82 & 0.39 & 0.38 & 0.67 & 0.33 & 0.92 & 0.28 & 0.39 & 0.94 & 21.00 & \textbf{\textcolor{blue}{23.00}} \\
 & GPT-4 & \textbf{\textcolor{red}{0.30}} & 0.88 & 0.57 & 0.35 & 0.88 & 0.42 & 0.87 & 0.46 & 0.42 & \textbf{\textcolor{red}{0.93}} & \textbf{\textcolor{red}{0.44}} & \textbf{\textcolor{red}{0.93}} & 0.27 & 0.38 & 0.92 & 7.00 & 10.00 \\
\cmidrule{1-19}
\multirow[t]{2}{*}{Element Aware + ICL + Stage 3} & Llama 3 8B & 0.19 & 0.86 & 0.44 & 0.39 & 0.85 & 0.25 & 0.82 & 0.42 & 0.42 & 0.35 & 0.31 & 0.91 & 0.35 & 0.40 & 0.92 & 19.00 & 19.00 \\
 & GPT-4 & 0.27 & 0.87 & 0.64 & 0.39 & 0.84 & 0.88 & 0.92 & 0.42 & 0.41 & 0.83 & 0.41 & 0.92 & 0.34 & 0.40 & 0.90 & \textbf{\textcolor{red}{2.00}} & 4.00 \\
\cmidrule{1-19}
\multirow[t]{2}{*}{Element Aware + ICL + Stage 2 (Deberta) + Stage 3} & Llama 3 8B & 0.25 & 0.87 & 0.58 & 0.37 & 0.71 & 0.17 & \textbf{\underline{\textcolor{blue}{0.81}}} & \textbf{\underline{\textcolor{red}{0.63}}} & 0.42 & 0.77 & 0.39 & 0.92 & 0.35 & 0.37 & 0.81 & 13.00 & 14.00 \\
 & GPT-4 & 0.25 & 0.87 & \textbf{\textcolor{red}{0.64}} & 0.39 & 0.88 & 0.42 & 0.86 & 0.49 & \textbf{\textcolor{red}{0.45}} & 0.92 & 0.41 & 0.92 & 0.34 & 0.40 & 0.89 & \textbf{\underline{\textcolor{red}{1.00}}} & \textbf{\underline{\textcolor{red}{1.00}}} \\
\cmidrule{1-19}
\multirow[t]{2}{*}{Element Aware + ICL + Stage 2 (PubMedBERT) + Stage 3} & Llama 3 8B & 0.27 & 0.87 & 0.56 & 0.37 & \textbf{\underline{\textcolor{blue}{0.62}}} & 0.29 & 0.81 & 0.54 & 0.42 & 0.59 & 0.39 & 0.92 & 0.36 & 0.37 & 0.81 & 11.00 & 8.00 \\
 & GPT-4 & 0.27 & 0.87 & 0.64 & 0.39 & 0.83 & \textbf{\textcolor{red}{0.90}} & \textbf{\textcolor{red}{0.92}} & 0.41 & 0.42 & 0.80 & 0.41 & 0.92 & 0.34 & 0.39 & 0.90 & 3.00 & \textbf{\textcolor{red}{2.00}} \\
\cmidrule{1-19}
\multirow[t]{2}{*}{Element Aware + ICL + Stage 2 (Reflection) + Stage 3} & Llama 3 8B & 0.24 & 0.87 & 0.60 & 0.39 & 0.64 & 0.27 & \textbf{\textcolor{blue}{0.81}} & \textbf{\textcolor{red}{0.58}} & 0.42 & 0.49 & 0.32 & \textbf{\textcolor{blue}{0.90}} & \textbf{\textcolor{red}{0.39}} & 0.37 & \textbf{\underline{\textcolor{blue}{0.65}}} & 15.00 & 11.00 \\
 & GPT-4 & 0.27 & 0.87 & \textbf{\underline{\textcolor{red}{0.65}}} & \textbf{\textcolor{red}{0.39}} & 0.84 & 0.88 & 0.92 & 0.42 & 0.42 & 0.80 & 0.40 & 0.92 & 0.34 & 0.39 & 0.88 & 5.00 & \textbf{\textcolor{red}{2.00}} \\
\bottomrule
\end{tabular}

%% file: algo_2_merge_missing_sentences.tex
In Algorithm \ref{alg:2}, ${\rm top}_K$ represents the thresholds for the number of missing information to merge, \({\rm PPL}_{\rm LM}(k^i_{\rm missing}, {\rm summ}_{\rm out},l)\) represents the perplexity of the text formed by inserting $i$-th missing information $k^i_{\rm missing}$ after $l$-th sentence of (updated) Stage 2 summary ${\rm summ}_{\rm out}$, and ${\rm insert}(k^i_{\rm missing}, {\rm summ_{\rm out}}, l^*)$ represents inserting $i$-th missing information $k^i_{\rm missing}$ after $l$-th sentence of (updated) Stage 2 summary ${\rm summ_{\rm out}}$.

\begin{algorithm}[H]
\caption{Merge Missing Information to Summary}
\label{alg:2}
\begin{algorithmic}[1]
  \STATE ${\rm summ_{out}} \gets $ \text{Stage 2 summary}
  \STATE $i \gets 0$
  \WHILE{$ i < {\rm top}_K$}
    \STATE $s \gets$ number of sentences in $\rm summ_{out}$
    \STATE ${\rm locs} := \{l \mid 0 \leq l \leq s\}$

    \STATE \(
      {l}^* := \operatorname*{argmin}_{l \in \rm locs} {\rm PPL}_{\rm LM}(k^i_{\rm missing}, {\rm summ}_{\rm out},l)
    \)
    
    \STATE ${\rm summ_{out}} \gets {\rm insert}(k^i_{\rm missing}, {\rm summ_{out}}, l^*)$
    \STATE $i \gets i + 1$
  \ENDWHILE
  \RETURN $\rm summ_{out}$
\end{algorithmic}
\end{algorithm}